\documentclass[12pt,reqno]{amsart}
\usepackage{graphicx,amssymb,amsmath}
\usepackage{amsaddr}
\usepackage[utf8]{inputenc}
\usepackage{epstopdf}
\usepackage{url,hyperref}	
\usepackage{bm}
\usepackage{color}
\usepackage{array}
\usepackage{enumerate}
\usepackage{doi}

\usepackage{theorem}
\usepackage[nodayofweek]{datetime}

\usepackage{natbib}
\setcitestyle{authoryear,open={(},close={)}}
\usepackage{hyperref}
\usepackage{etoolbox}
\usepackage{chemformula}
\usepackage{gensymb}

\newcommand*{\G}{{\boldsymbol{G}}}

\renewcommand*{\Xi}{{\boldsymbol{\xi}}}

\newcommand*{\R}{{\mathbb{R}}}

\renewcommand*{\u}{{\boldsymbol{u}}}
\newcommand*{\bzeta}{\boldsymbol{\zeta}}
\newcommand*{\y}{{\boldsymbol{y}}}
\newcommand*{\yobs}{{\boldsymbol{y}}^{\rm{o}}}
\newcommand*{\zetaobs}{{\boldsymbol{\zeta}}^{\rm{o}}}

\newcommand*{\x}{{\boldsymbol{x}}}
\newcommand*{\Win}{{\boldsymbol{W}_{\rm{in}}}}
\newcommand*{\bin}{{\boldsymbol{b}_{\rm{in}}}}
\newcommand*{\Robs}{\bm{\mathrm{\Gamma}}}
\newcommand*{\W}{{\boldsymbol{W}}}
\newcommand*{\X}{{\boldsymbol{X}}}
\newcommand*{\Ntrain}{N}
\newcommand*{\Id}{\boldsymbol{I}}
\newcommand*{\Res}{{\boldsymbol{\Phi}}}

\newcommand*{\w}{{\boldsymbol{w}}}
\newcommand*{\K}{{\boldsymbol{K}}}
\renewcommand*{\H}{{\boldsymbol{H}}}

\newcommand*{\Pf}{{\boldsymbol{P}}^{\rm f}}

\newcommand*{\Z}{{\boldsymbol{Z}}}

\newcommand{\tauf}{\tau_f}

\usepackage[ruled,vlined,boxed,commentsnumbered]{algorithm2e}

\title{Combining machine learning and data assimilation to forecast dynamical systems from noisy partial observations}

\author[G. A. Gottwald]{Georg A. Gottwald}
\address[G. A. Gottwald]
{School of Mathematics and Statistics \\
 University of Sydney \\
 NSW 2006 \\
 Australia
}
\author[S. Reich]{Sebastian Reich}
\address[S. Reich]
{Institute of Mathematics\\
 University of Potsdam \\
Germany
}
\email[G. A. Gottwald]{georg.gottwald@sydney.edu.au} 
\email[S. Reich]{sebastian.reich@uni-potsdam.de}

\begin{document}

\maketitle


\begin{abstract}
We present a supervised learning method to learn the propagator map of a dynamical system from partial and noisy observations. In our computationally cheap and easy-to-implement framework a neural network consisting of random feature maps is trained sequentially by incoming observations within a data assimilation procedure. By employing Takens' embedding theorem, the network is trained on delay coordinates. We show that the combination of random feature maps and data assimilation, called RAFDA, outperforms standard random feature maps for which the dynamics is learned using batch data.
\end{abstract}


{\bf{Determining computationally inexpensive surrogate maps, trained on partial and noisy observations, is of utmost significance. Such surrogate maps would allow for long-time simulation of dynamical systems, the governing equations of which are unknown. Similarly, they would allow for long-time integration of complex high-dimensional dynamical systems with known underlying equations where one is however only interested in the dynamics of a select subset of resolved variables. Surrogate maps should produce trajectories which are consistent with the dynamical system even after long time integration. We consider here the situation when such systems are accessible only by partial observations of the resolved variables and where these observations are contaminated by measurement noise. By formulating a machine learning framework in delay coordinate space, we show that a partial and noisy training data set can be used to learn the dynamics in the reconstructed phase space. The learned surrogate map provides a computationally cheap forecast model for single forecasts up to several Lyapunov times as well as for dynamically consistent long-time simulations.}}

\section{Introduction}
Recent years have seen an increased interest in data-driven methods with the aim to develop cheap surrogate models to perform forecasting of dynamical systems. A main driver behind this endeavour is the potential saving of computational running time required for simulation. This is particularly important for stiff multi-scale systems for which the fastest time-scale puts strong restrictions on the time step which an be employed \citep{HanEtAl18,HanEtAl19,RaissiEtAl18,RaissiEtAl19}. Promising applications for such surrogate models come from atmospheric and ocean dynamics and from climate dynamics \citep{SchneiderEtAl17,DuebenBauer18,RaspEtAl18,GagneEtAl19,BoltonZanna19,BCBB20b,Nadiga21,ClearyEtAl21}. 
 
A particularly simple and efficient machine learning architecture are random feature maps \citep{RahimiRecht08, RahimiRecht08b,Bach17a,Bach17b,SunGilbertTewari19}. Random feature maps provide a representation of the surrogate propagator for a dynamical system by a linear combination of randomly generated high-dimensional nonlinear functions of the input data. The training of random feature map networks only requires linear least-square regression and it was proven rigorously that random feature maps enjoy the so called universal approximation property which states that they can approximate any continuous function arbitrarily close \citep{ParkSandberg91,Cybenko89,Barron93}. The framework of random feature maps was extended to include internal dynamics in so called echo-state networks and reservoir computers with remarkable success in forecasting dynamical systems \citep{MaassEtAl02,Jaeger02,JaegerHaas04,PathakEtAl18,JuenglingEtAl19,AlgarEtAl19,Nadiga21,Bollt21,GauthierEtAl21,PlattEtAl21}. 

These linear regression based training methods assume model errors rather than measurement errors in the data \citep{GottwaldReich21}. However, the skill of random feature maps in providing a reliable surrogate forecast model can be severely impeded when the data used for training are contaminated by measurement noise. 
In previous work \citet{GottwaldReich21} showed that by combining random feature maps with ensemble-based sequential data assimilation, in a framework coined \emph{RAndom Feature map Data Assimilation} (RAFDA), the noise can be effectively controlled leading to remarkable forecast skills for noisy observations. Moreover, RAFDA was shown to be able to provide model closure in multi-scale systems allowing for subgridscale parametrization as well as to provide reliable ensembles for probabilistic forecasting.  In RAFDA the parameters of the random feature maps are learned sequentially with incoming observations in conjunction with the analysis of the state variables using an ensemble Kalman filter (EnKF) \citep{Evensen}. RAFDA thus combines a powerful approximation tool in the form of random feature maps with an advanced sequential and derivative-free data assimilation technique, which is particularly well suited for high-dimensional state and parameter estimation problems.

Several alternative combinations of data assimilation and machine learning techniques have recently been explored.  Recently proposed methodologies include purely minimization-based approaches \citep{BBCB19}, approaches which combine ensemble-based sequential state estimation with minimization-based parameter estimation \citep{BCBB20a,BBCB20}, and purely ensemble-based sequential joined state and parameter estimation \citep{BocquetEtAl_FDS2021}. Furthermore, an intimate formal equivalence between certain neural network architectures and data assimilation has been noted by \citet{ARS18}. Finally, sequential data assimilation has also been extended to echo-state networks and reservoir computers \citep{TomizawaSawada20,WiknerEtAl21}. 

In this work, we extend RAFDA to the situation when only partial observations are available.
In this case, the propagator map, that updates the system in the partially observed subspace only, is non-Markovian and knowledge of the past history is required. This problem was recently addressed within a machine learning context using appropriate model closures \citep{LevineStuart21}. To account for the non-Markovian nature of the propagator map, we follow here a different path instead and employ phase-space reconstruction and Takens' embedding theorem \citep{Takens81}. Takens' embedding theorem has been successfully used in data assimilation before where a forecast model was non-parametrically constructed using analogs \citep{HamiltonEtAl16}. Recently reservoir computers, which may be viewed as providing an embedding of the dynamics in the reservoir space \citep{LuEtAl17,HartEtAl20}, were used to reconstruct dynamics in the reconstructed phase space using delay coordinates and linear regression for parameter estimation \citep{NakaiSaiki21}. Here we use delay coordinates in combination with sequential data assimilation to learn a surrogate forecast model directly in the reconstructed phase-space from noisy and partial observations.

While we follow a non-parametric or model-free approach in this paper, another promising direction is to combine machine learning and knowledge-based modelling components  \citep{PathakEtAl_CHAOS2018,GagneEtAl19,BoltonZanna19,WiknerEtAl_CHAOS2020,BCBB20b,WiknerEtAl21}. We note that RAFDA also allows for such extensions \citep{GottwaldReich21}.

The paper is organised as follows. In Section~\ref{sec:RAFDA} we develop our RAFDA methodology for partial observations. In Section~\ref{sec:L63} we show how RAFDA performs on the Lorenz 63 system and that RAFDA exhibits significantly increased forecast skill when compared to standard random feature maps trained on delay coordinates using linear regression. We close in Section~\ref{sec:discussion} with a discussion and an outlook.


\section{Random feature map and data assimilation (RAFDA)}
\label{sec:RAFDA}
Consider a $D$-dimensional dynamical system $\dot {\u} = \mathcal{\boldsymbol{F}}(\u)$ which is accessed at equidistant times $t_n = n\Delta t$ of interval length $\Delta t>0$, $n \ge 0$, by partial noisy observations
\begin{align} 
\label{eq:obs}
\yobs_n = \boldsymbol{G} \u_n + \Robs^{1/2}\,  \boldsymbol{\eta}_n
\end{align}
with  $\u_n = \u(t_n)$, observation operator $\boldsymbol{G}:\R^D\to \R^d$, measurement error covariance matrix $\Robs\in \R^{d\times d}$ and $d$-dimensional independent and normally distributed noise $\boldsymbol{\eta}_n$, that is, $\boldsymbol{\eta}_n \sim {\mathcal N}({\bf 0},{\bf I})$. 

The aim we set out to pursue is the following: using noisy observations $\yobs_j$ for $0\le j\le n$ find an approximation to the propagator map which maps previous observations (or a suitable subset of them) to the unobserved variable $\y_{n+1}$ at future time $t_{n+1}$. 
In the Markovian case $d=D$ with $\boldsymbol{G}=\Id$ the propagator map is only a function of the observation at the current time $t_n$. This case was studied in our previous work \citep{GottwaldReich21}. Here the focus is on partial observations with $d<D$ for which the propagator map is non-Markovian. Our method judiciously incorporates three separate methods which we discuss in the following three subsections: 1.) Takens embedding and delay coordinates to deal with the aspect of having only access to partial observations and the resulting non-Markovian propagator map, 2.) random feature maps which form our approximation for the propagator map and 3.) data assimilation and EnKFs to control the observation noise and to determine the parameters of the random feature maps leading to the desired surrogate model.


\subsection{Delay coordinates}
\label{sec:delay}
When only partial observations are available Takens' embedding theorem allows to represent the dynamics faithfully by means of phase space reconstruction and delay vectors \citep{Takens81,SauerEtAl91,SauerYorke93}. Takens' embedding theorem assumes noise-free observations, but has been successfully applied to noise-contaminated observations \citep{KantzSchreiber,Schreiber91,Small,CasdagliEtAl91,SchreiberKantz95}. 

We describe here the key steps for scalar-valued observations, that is, $d = 1$ in (\ref{eq:obs}). 
Hence, given a time-series $y_n^{\rm o}$, define the $m$-dimensional delay vectors
\begin{align} \label{eq:delay_coordinates}
\zetaobs_n &=\left(y_n^{\rm{o}},y_{n+\tau}^{\rm{o}},y_{n+2\tau}^{\rm{o}},...,y_{n+(m-1)\tau}^{\rm{o}}\right)^{\rm T}
\end{align}
for $n=0,\ldots,N$, with integer delay time $\tau> 0$. If the underlying dynamics lies inside an $D$-dimensional phase space, then the delay reconstruction map $y_n^{\rm o} \to \bzeta_n^{\rm o}$ is an embedding provided the embedding dimension is sufficiently large with $m > 2\,D_{\rm{box}}$ where $D_{\rm{box}}$ denotes the fractal box-counting dimension of the attractor \citep{SauerEtAl91,SauerYorke93}. 
Under these conditions, the reconstructed dynamics $\zetaobs_0,\zetaobs_1,\zetaobs_2,\dots$ in $\R^m$ faithfully represents the underlying dynamics. For finite time series, the choice of the embedding dimension $m$ and the delay time $\tau$ are crucial and need to be carefully chosen. See for example \citet{KantzSchreiber,Schreiber91,CasdagliEtAl91,SchreiberKantz95,Small} for a detailed discussion. Specifically, if the embedding dimension is chosen too small, then the delay reconstruction map does not represent the underlying dynamics, if it is chosen too large, the reconstruction involves redundancy reducing the length of the delay vector time series $\zetaobs_n$. We choose the embedding dimension $m$ using the false nearest neighbour algorithm \citep{KennelEtAl92}. The embedding dimension is chosen as the smallest value $m$ such that the number of false nearest neighbours is less than $10\%$. False nearest neighbours (at dimension $m$) are defined as those points $\zetaobs_n$ for which the Euclidean distance to their nearest neighbour changes upon increasing the embedding dimension to $m+1$ by a factor of $10$ (relative to their smallest Euclidean distance). An appropriate delay time $\tau$ is determined by determining the time for which the average mutual information has its first zero-crossing \citep{FraserSwinney86}. We use the command {\texttt{phaseSpaceReconstruction}} from \citet{Matlab} which implements these algorithms to determine the embedding dimension $m$, the delay time $\tau$ as well as generating the delay vectors. 

The extension of the delay vectors (\ref{eq:delay_coordinates}) to multivariate observations (\ref{eq:obs}) is straightforward and leads to delay vectors of dimension $D_\zeta = dm$.



\subsection{Random feature maps}
\label{sec:rfm}
In this subsection we describe the traditional random feature map network architecture \citep{RahimiRecht08, RahimiRecht08b,Bach17a,Bach17b,SunGilbertTewari19}. This is a particularly easy-to-implement network architecture in which the input is given as a linear combination of randomly sampled nonlinear functions of the input signal, and the coefficients of this linear combination are then learned from the whole available training data set via linear ridge regression. Our RAFDA extension, as described further below, instead determines the coefficients sequentially within a data assimilation procedure.

Applied to our setting, the delay coordinates $\bzeta \in \R^{D_\zeta \times 1}$ are first linearly mapped by a random but fixed linear map into a high-dimensional subspace of $\R^{D_r}$ with $D_r\gg D_\zeta$ and then nonlinearly transformed by feature maps $\boldsymbol{\phi}$ as
\begin{align}
\boldsymbol{\phi} (\bzeta) = \tanh(\Win \bzeta + \bin) \in \mathbb{R}^{D_r \times 1}
\label{eq:r}
\end{align}
with weight matrix 
\begin{align*}
\Win = (\w_{{\rm in},1},\ldots,\w_{{\rm in},D_r})^{\rm T} \in \mathbb{R}^{D_r\times D_\zeta}
\end{align*}
and a bias
\begin{align*}
\bin = (b_{{\rm in},1},\ldots,b_{{\rm in},D_r})^{\rm T} \in \mathbb{R}^{D_r \times 1}.
\end{align*}
The weight matrix and the bias are chosen randomly and independently of the observed delay coordinates $\bzeta_n^{\rm o}$, $n=0,\ldots,N$,
according to the distributions $p(\w_{\rm in})$ and $p(b_{\rm in})$, respectively. We choose here
\begin{align}
(\Win)_{ij} \sim \mathcal{U}[-w,w] \qquad {\rm{and}} \qquad (\bin)_i \sim \mathcal{U}[-b,b] .
\label{eq:wb}
\end{align}
The choice of the hyperparameters $w>0$ and $b>0$ is system-dependent and needs to ensure that the observations cover the nonlinear domain of the $\tanh$-function. The reader is referred to \citet{GottwaldReich21} for further details. Note that the hyperparameters $\Win$ and $\bin$ are kept fixed once drawn and are not learned. This restriction is made for computational simplicity and can be relaxed following \citet{EEtAl20,RotskoffVandenEijnden18}.

An approximation of the propagator map in the delay coordinates $\boldsymbol{\zeta}\in R^{D_\zeta}$ is then provided by
\begin{align}
\Psi_{\rm S}(\bzeta) = \W \boldsymbol{\phi}(\bzeta),
\label{eq:Wr}
\end{align}
where the matrix $\W \in \mathbb{R}^{D_\zeta \times D_r}$ maps back into the delay coordinate vector space to approximate the delay vector at the next time step, and is to be learned from the noisy delay vectors $\zetaobs_n$, $n=0,\ldots,N$. More specifically, the weights $\W$ should be chosen such that 
\begin{align}
    \zetaobs_{n} &\approx
    \Psi_{\rm S}(\zetaobs_{n-1}),
\end{align}
which is achieved by minimizing the regularized cost function
\begin{align} \label{eq:RR}
\mathcal{L}(\W) &= \frac{1}{2} \sum_{n=1}^{\Ntrain}
\|\zetaobs_{n} -\Psi_{\rm S}(\zetaobs_{n-1})\|^2 + \frac{\beta}{2}
\|\W\|^2_{\rm F}
\nonumber
\\ &=
\frac{1}{2} \| \Z^{\rm o}-\W \Res \|_{\rm F}^2 + \frac{\beta}{2} \|\W\|^2_{\rm F},
\end{align}
where $\|\mathbf{A}\|_F$ denotes the Frobenius norm of a matrix $\mathbf{A}$,  $\Z^{\rm o}\in \R^{D_\zeta \times \Ntrain}$ is the matrix with columns $\zetaobs_{n}$, $n=1,\ldots,\Ntrain$, and $\Res\in \R^{D_r\times \Ntrain}$  is the matrix with columns 
\begin{equation} \label{eq:feature_map_r}
{\boldsymbol{\phi}}_{n} = \boldsymbol{\phi}(\zetaobs_{n-1}),
\end{equation}
$n=1,\ldots,\Ntrain$. The parameter $\beta > 0$ is used for regularization. The solution to the minimization problem for (\ref{eq:RR}) can be explicitly determined as
\begin{align}
\W_{\rm LR} = \Z^{\rm o}\Res^{\rm T} \left( \Res \Res^{\rm T}+ \beta {\bf I} \right)^{-1} \,,
\label{eq:WLR}
\end{align}
and uses all available training data $\Z^{\rm o}$ at once. In \citet{GottwaldReich21} it was shown that standard random feature maps have difficulty dealing with noisy observations, and instead it was proposed to learn the output weights $\W$ sequentially within a data assimilation procedure which is described in the next section.


\subsection{Data assimilation}
\label{sec:DA}

RAFDA uses a combined parameter and state estimation within a data assimilation procedure. The main idea is to use the surrogate propagator (\ref{eq:Wr}) consisting of the random feature maps as the forecast model within an EnKF and to estimate sequentially the weight matrix $\W$. Concretely, we consider the forecast model for given weight matrix $\W_{n-1}^{\rm a}$ and given delay coordinates $\bzeta_{n-1}^{\rm a}$ as
\begin{subequations} 
\label{eq:forecast}
\begin{align} 
\label{eq:forecast_a}
\bzeta^{\rm f}_{n} &= \W_{n-1}^{\rm a} \,\boldsymbol{\phi}(\bzeta_{n-1}^{\rm a})\\
\W^{\rm f}_{n} &= \W^{\rm a}_{n-1},
\end{align}
\end{subequations}
where the superscript f denotes the forecast and the superscript a denotes the analysis defined below. 

To incorporate the parameter estimation into an EnKF analysis step, we consider an augmented state space  $\x  = (\bzeta^{\rm T},\w^{\rm T})^{\rm T} \in \R^{D_x\times 1}$ with $D_x = D_\zeta(1 + D_r)$. Here $\w \in \R^{D_\zeta D_r \times 1}$ is the vector consisting of all matrix elements of the weight matrix $\W$ with its entries defined block-wise, that is, $w_{1:D_r}=(W_{11},\dots,W_{1D_r})^{\rm T}$, $w_{D_r+1:2D_r}=(W_{21},\dots,W_{2D_r})^{\rm T}$ and so forth. 

We now also treat $\x^{\rm f}_n$ and $\x^{\rm a}_n$, $n\ge 0$ as random variables. Assuming a Gaussian distribution for $\x_{n+1}^{\rm f}$, the analysis step for the mean $\overline \x^{\rm a}_{n}$ is given by
\begin{align}
\overline \x^{\rm a}_{n} = \overline \x^{\rm f}_{n} - \K_{n} (\H \overline \x^{\rm f}_{n}-\zetaobs_{n})
\label{eq:KF1}
\end{align}
with the observation matrix $\H \in \mathbb{R}^{D_\zeta \times D_x}$ defined by $\H \x = \bzeta$ and the measurement error covariance matrix of the delay vectors given by ${\bf \Gamma}^{\rm dc} = {\bf I} \otimes \Robs \in \R^{D_\zeta \times D_\zeta}$ (compare (\ref{eq:obs})). Here $\boldsymbol{A} \otimes \boldsymbol{B}$ denotes the Kronecker product of two matrices. The Kalman gain matrix $\K_n$ is given by
\begin{align} \label{eq:Kalman_gain}
\K_{n} = \Pf_{n} \H^{\rm T}\left(\H \Pf_{n}\H^{\rm T}+{\bf \Gamma}^{\rm dc}\right)^{-1}
\end{align}
with forecast covariance matrix 
\begin{align}
\Pf_{n} &=  \langle \hat \x^{\rm f}_{n} \otimes \hat \x_{n}^{\rm f} \rangle 
\nonumber
\\
&= 
\left( \begin{array}{cc}
  \langle\hat \bzeta^{\rm f}_{n} \otimes \hat\bzeta^{\rm f}_{n}\rangle  &  \langle\hat\bzeta^{\rm f}_{n}\otimes \hat\w^{\rm f}_{n}\rangle\\
  \langle\hat\w^{\rm f}_{n}\otimes \hat\bzeta^{\rm f}_{n}\rangle  &  \langle\hat\w^{\rm f}_{n} \otimes \hat\w^{\rm f}_{n}\rangle\\
\end{array} \right).
\label{eq:Pf}
\end{align}
The angular bracket denotes the expectation value and the hat denotes the perturbation of $\x_{n}^{\rm f}$ from its mean $\overline{\x}_{n}^{\rm f} = \langle \x_{n}^{\rm f}\rangle$, as in 
\begin{align}
\hat \bzeta^{\rm f}_{n} = \bzeta_{n}^{\rm f} - \overline{\bzeta}_{n}^{\rm f}.
\end{align}
Since $\H\x=\bzeta$, we can separate the state and parameter update of the Kalman analysis step (\ref{eq:KF1}) as
\begin{subequations}
\begin{align}
\overline \bzeta^{\rm a}_{n} &= \overline \bzeta^{\rm f}_{n} -  \Pf_{\bzeta\bzeta} \left( \Pf_{\bzeta\bzeta}+{\bf \Gamma}^{\rm dc} \right)^{-1}\Delta \boldsymbol{I}_{n}\\
\overline \w^{\rm a}_{n} &= \overline \w^{\rm f}_{n} -  \Pf_{\w\bzeta} \left( \Pf_{\bzeta\bzeta}+{\bf \Gamma}^{\rm dc}\right)^{-1}\Delta \boldsymbol{I}_{n}
\end{align}
\label{eq:KF2}
\end{subequations}
with innovation
\begin{align}
    \Delta \boldsymbol{I}_{n} := \overline \bzeta_{n}^{\rm f}-\zetaobs_{n}
\end{align}
and covariance matrices
$ \Pf_{\bzeta\bzeta} =  \langle\hat \bzeta^{\rm f}_{n} \otimes \hat\bzeta^{\rm f}_{n}\rangle $ and $\Pf_{\w\bzeta}= 
\langle\hat\w^{\rm f}_{n}\otimes \hat\bzeta^{\rm f}_{n}\rangle$. Note that the delay vectors $\zetaobs_n$ are correlated in time as after each $\tau$ steps $(m-1)d$ components reappear. This implies that the setting described above is not strictly Bayesian. However, if the delay time $\tau$ is sufficiently large the dynamics of the combined forecast-analysis dynamical system will have sufficiently decorrelated and for practical purposes we can treat the observations as independent.

To implement the Kalman analysis step, we employ a stochastic EnKF \citep{BurgersEtAl98,Evensen}. This allows to estimate the forecast covariances adapted to the dynamics and has advantages for the nonlinear forward model (\ref{eq:forecast}) and the non-Gaussian augmented state variables. Consider an ensemble of states $\X \in \mathbb{R}^{D_x\times M}$ consisting of $M$ members $\x^{(i)}\in \mathbb{R}^{D_x \times 1}$, $i=1,\ldots,M$, that is,
\begin{equation}
\X=\left[ \x^{(1)},\x^{(2)},\dots,\x^{(M)} \right],
\end{equation}
with empirical mean
\begin{equation}
\overline{\x} = \frac{1}{M} \sum_{i=1}^M  \x^{(i)},
\end{equation}
and associated matrix of ensemble deviations
\begin{equation}
\hat \X=\left[ \x^{(1)}-\overline{\x},\x^{(2)}-\overline{\x},\dots,\x^{(M)}-\overline{\x} \right].
\end{equation}
Ensembles for the forecast are denoted again by superscript f and those for the analysis by superscript a. In the forecast step each ensemble member is propagated individually using (\ref{eq:forecast}), updating the previous analysis ensemble $\X_{n-1}^{\rm a}$ to the next forecast ensemble $\X_{n}^{\rm f}$. The forecast covariance matrix (\ref{eq:Pf}) used in the analysis step (\ref{eq:KF1}) can be estimated as a Monte-Carlo approximation from the forecast ensemble deviation matrix $\hat \X_{n}^{\rm f}$ via
\begin{align} \label{eq:P_empirical}
\Pf_{n} = 
\frac{1}{M-1}\hat \X_{n}^{\rm f}\,(\hat\X^{\rm f}_{n})^{\rm T}
\in \mathbb{R}^{D_x\times D_x} .
\end{align}

In the stochastic ensemble Kalman filter observations $\zetaobs_{n}$ receive a stochastic perturbation $\boldsymbol{\eta}_{n}^{(i)}\in \R^{D_\zeta \times 1}$, $i=1,\ldots,M$, drawn independently from the Gaussian observational noise distribution $\mathcal{N}({\bf 0},{\bf \Gamma}^{\rm dc})$. The associated ensemble of perturbed observations $\Z^{\rm p}_{n} \in \mathbb{R}^{D_\zeta \times M}$ is given by
\begin{align}
\Z^{\rm p}_{n} &= \left[ \zetaobs_{n} -  \boldsymbol{\eta}_{n}^{(1)},\zetaobs_{n} -\boldsymbol{\eta}_{n}^{(2)},
\ldots,
\zetaobs_{n} -  \boldsymbol{\eta}_{n}^{(M)} \right].
\end{align}
The EnKF analysis update step is then given by
\begin{align} 
\label{eq:EnKF}
\X_{n}^{\rm a} = \X_{n}^{\rm f} - \boldsymbol{K}_{n} \Delta {\boldsymbol{I}}_{n},
\end{align}
with the Kalman gain defined by (\ref{eq:Kalman_gain}) using (\ref{eq:P_empirical}) and stochastic innovation
\begin{align} 
\Delta {\boldsymbol{I}}_{n}=  \H\X^{\rm f}_{n}-\Z^{\rm p}_{n}.
\end{align}
To mitigate against finite ensemble size effects covariance inflation with $\Pf_n \to \alpha \Pf_n$ is typically introduced with $\alpha>1$ \citep{AndersonAnderson99}. Such multiplicative inflation preserves the ensemble mean but increases the forecast error covariance. In \citet{GottwaldReich21}, which considers the fully observed case with $D=d$, localisation was employed by only considering covariances between a component of the state variable and its associated block in the parameter $\w$. This localisation was found here not to be advantageous; we believe that this is due to the fact that each component of the delay vectors appears in all components at different times. We therefore do not perform any localisation in the numerical simulations presented below.

The initial ensemble $\X_0^{\rm a}$ is initialized as $\bzeta_0^{\rm a}\sim {\mathcal N}(\zetaobs_0,{\bf \Gamma}^{\rm dc})$ and $\w_0^{\rm a} \sim {\mathcal N}(\w_{\rm LR},\gamma \,\mathbf{I})$ where $\w_{\rm LR}$ is the vectorial form of the solution $\W_{\rm LR}$ to the ridge regression formulation (\ref{eq:RR})
and $\gamma >0$ is a parameter specifying the spread of the initial ensemble.

We remark that the EnKF does not minimize the cost function (\ref{eq:RR}) but rather approximates the posterior distribution in the weights $\W$ given the observed delay vectors $\zetaobs_n$, $n=0,\ldots,N$, under the assumed measurement model (\ref{eq:obs}) and vanishing model errors in the propagator (\ref{eq:Wr}). Including stochastic model errors into RAFDA would be straightforward.

The ensemble forecast step defined by (\ref{eq:forecast}), together with the EnKF analysis step (\ref{eq:EnKF}) constitute 
our combined RAFDA method. We run RAFDA for a single long training data set of length $N$. The approximation of the propagator map is given by the random feature model (\ref{eq:Wr}) where the weight matrix $\W$ is given by the ensemble mean of the weight matrix  $\W_N^{\rm a}$ at final training time $t_N = \Delta t N$ and is denoted by $\W_{\rm RAFDA}$. 

We summarize our RAFDA method in Algorithm \ref{algo:RAFDA}.

\IncMargin{1em}
\begin{algorithm}
\SetKwData{Left}{left}\SetKwData{This}{this}\SetKwData{Uropagator}{up}
\SetKwFunction{Union}{Union}\SetKwFunction{FindCompress}{FindCompress}
\SetKwInOut{Input}{input}\SetKwInOut{Output}{output}
\SetKwInOut{InputData}{input data}
\SetKwInOut{InputPara}{parameters}
\SetKwInOut{Construct}{construct}
\SetKwInOut{Compute}{compute}

\InputData{time series $\yobs_n$, $n=0,\dots,N$}
\InputPara{random feature maps: dimension $D_r$, internal parameters $\Win\in \R^{D_r\times D_\zeta}$, $\bin\in\R^{D_r \times 1}$\\
EnKF: ensemble size $M$, \\measurement error covariance\\ ${\bf \Gamma}^{\rm dc} = {\bf I} \otimes\Robs$, inflation $\alpha$, initial\\ ensemble parameters $(\w_{\rm LR},\gamma)$}
\BlankLine
\textsc{perform the following:}\\
\texttt{delay coordinate embedding:}\\
$\quad\,$ form delay vectors $\zetaobs_n \in \mathbb{R}^{D_\zeta \times 1}$;\\
\texttt{initializing ensemble:}\\
$\quad\,$  set $\X_0^{\rm a}$ with  members drawn according to\\ $\quad\,$  $\bzeta_0^a \sim {\mathcal N}(\zetaobs_0,{\bf \Gamma}^{\rm dc})$ and $\w_0^a \sim {\mathcal N}(\w_{\rm LR},\gamma\, \mathbf{I})$\;
\For{$n=1 :  N$}{
 \texttt{forecast $\X_{n-1}^{\rm a} \to \X_{n}^{\rm f}$:} each ensemble member is propagated according to\\
 $\qquad$           $\bzeta^{\rm f}_{n} = \W_{n-1}^{\rm a} \,\phi(\bzeta^{\rm a}_{n-1})$\;
 $\qquad$           $\W^{\rm f}_{n} = \W^{\rm a}_{n-1}$\;
  \texttt{data assimilation analysis update:} \\
$\qquad$ inflation: $\Pf_n\leftarrow \alpha \Pf_n$\\
$\qquad$  $\X_{n}^{\rm a} = \X_{n}^{\rm f} - \boldsymbol{K}_{n} \left(\H\X^{\rm f}_{n}-\Z^{\rm p}_{n}\right)$\;
}
\Output{$\W_{\rm RAFDA} = {\text{ensemble average of }} \W^{\rm{a}}_N$}
\caption{Random Feature Map DataAssimilation (RAFDA)}
\label{algo:RAFDA}
\end{algorithm}\DecMargin{1em}


\section{Numerical results}
\label{sec:L63}
We consider the Lorenz-63 system \citep{Lorenz63}
\begin{subequations}
\label{eq:L63}
\begin{align}
\dot y &= 10(y-x) \\
\dot x &= 28 x - y -xz \\
\dot z &= -\frac{8}{3}z + xy
\end{align}
\end{subequations}
with $\u=(x,y,z)^{\rm T} \in \mathbb{R}^3$. Observations are only available for $x$, i.e.~$\G = (1\; 0\; 0)$ and $d=1$ in (\ref{eq:obs}), and are taken every $\Delta t = 0.02$ time units. We employ observational noise with measurement error covariance $\Robs=\eta {\bf I}$ and use $\eta=0.2$ unless stated otherwise (compare (\ref{eq:obs})). We ensure that the dynamics evolves on the attractor by discarding an initial transient of $40$ model time units. The optimal embedding dimension and delay time for these parameters are estimated as $m=3$ and $\tau=10$. In the following times are measured in units of the Lyapunov time $t\lambda_{\rm{max}}$ with the maximal Lyapunov exponent $\lambda_{\rm{max}}=0.91$. 

We use a reservoir of size $D_r=300$ with internal parameters $w=0.005$ and $b=4$, and an ensemble size of $M=300$ for the ensemble Kalman filter. The regularization parameter is set to $\beta=2\times 10^{-5}$ 
and the inflation parameter to $\alpha = 1 +0.01\Delta t=1.0002$. We employ a training set of length $N=4,000$.

To test the propensity of RAFDA to learn a surrogate model (\ref{eq:Wr}) we generate a validation data set $\bzeta_{\rm valid} (t_n)$, $n\ge 0$ sampled with the same rate as the training data set with $\Delta t = 0.02$ from the $x$-component of the Lorenz-63 system (\ref{eq:L63}). To quantify the forecast skill we measure the forecast time $\tauf$, defined as the largest time such that the relative forecast error $\mathcal{E}(t_n)=||\bzeta_{\rm valid}(t_n)-\bzeta_n ||^2/||\bzeta_{\rm valid}(t_n)||^2\le \theta$, where the $\bzeta_n$ are generated by the learned surrogate model. We choose here $\theta=40$. 

The forecast skill depends crucially on the choice of the randomly chosen internal parameters $(\Win,\bin)$ as well as on the training and validation data set. We therefore report on the mean behaviour over $500$ realisations, differing in the training and validation data set, in the random draws of the internal parameters $(\Win,\bin)$ and in the initial ensembles for RAFDA. Each training and validation data set is generated from randomly drawn initial conditions which are evolved independently over $40$ model time units to ensure that the dynamics has settled on the attractor.

\begin{figure}[tbp]
\centering
\includegraphics[width=0.8\columnwidth]{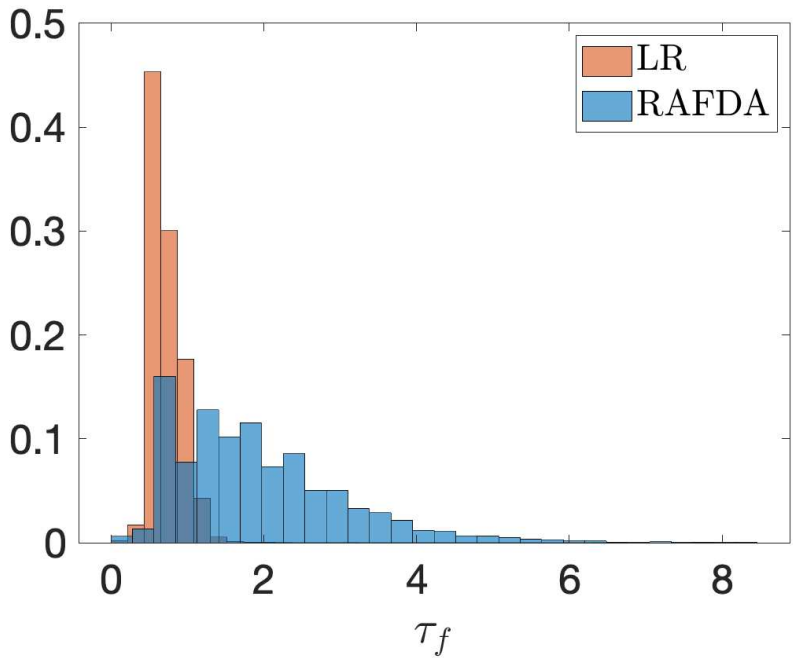}
\caption{Empirical histogram of forecast times $\tauf$ in units of the Lyapunov time for noise contaminated observations with $\eta=0.2$. Results are shown for standard LR and for RAFDA.}
\label{fig:L63_hist}
\end{figure}

\begin{figure}[tbp]
\centering
\includegraphics[width=0.8\columnwidth]{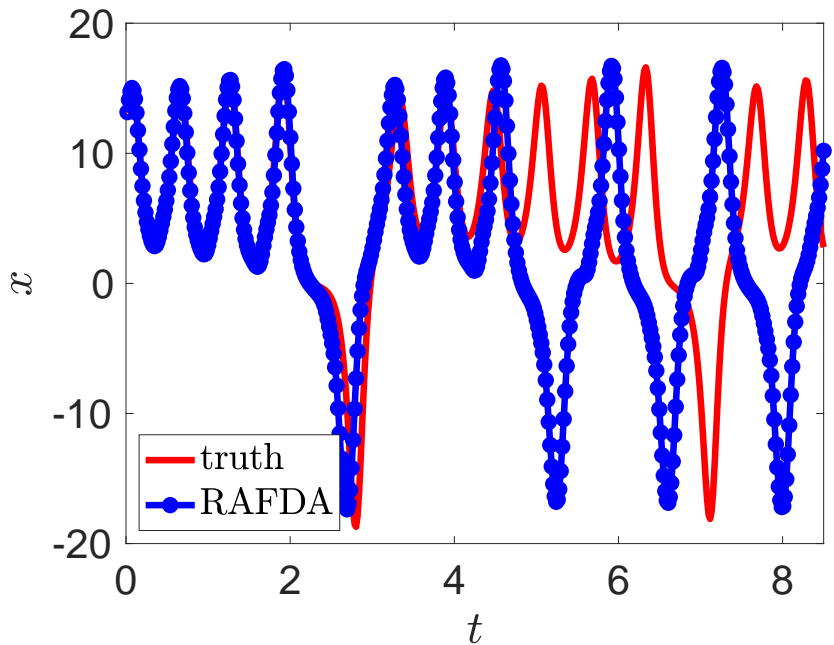}
\caption{Comparison of a representative validation time series (truth) and the corresponding RAFDA forecast, initialized with the first three data points from the truth. The measured forecast time of $\tauf=2.6$ is close to the mean forecast time.}
\label{fig:L63_validation}
\end{figure}

\begin{figure}[tbp]
\centering
\includegraphics[width=0.8\columnwidth]{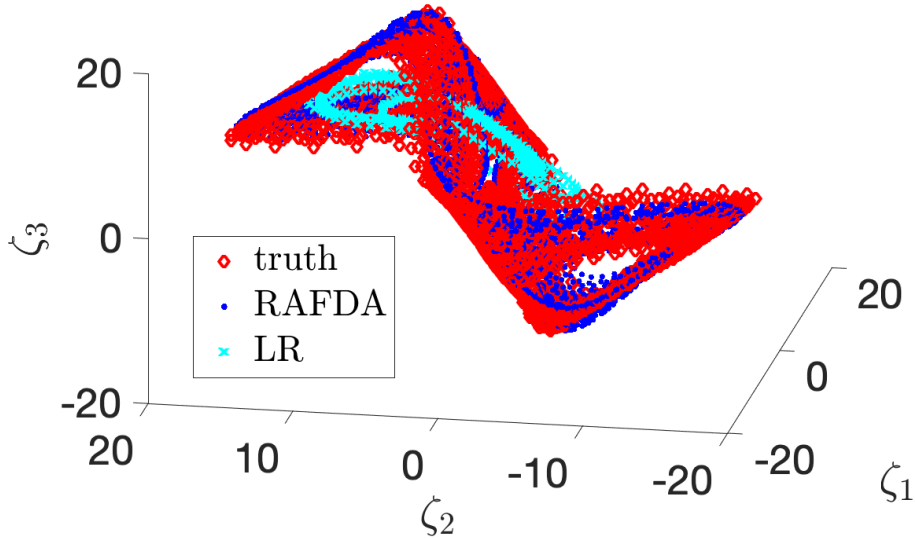}
\caption{Attractor in the reconstructed phase space spanned by the delay-coordinates. The truth is depicted by diamonds (online red), RAFDA by filled circles (online blue) and LR by crosses (online cyan).}
\label{fig:L63_attractor}
\end{figure}

In the following, we compare RAFDA with standard random feature maps using linear regression (LR). Figure~\ref{fig:L63_hist} shows the empirical histogram for both RAFDA and LR when the surrogate model (\ref{eq:Wr}) is trained on data contaminated with noise of strength $\eta=0.2$. The mean forecast time obtained by RAFDA is $\tauf=2.12$ and is roughly three times larger than the forecast time obtained using standard random feature maps which yields $\tauf=0.77$. Furthermore, it is seen that the distribution of forecast times for RAFDA is heavily skewed towards larger forecast times. 
As expected, these forecast times are smaller than for the fully observed case with $\G={\bf I}$, as considered in \citet{GottwaldReich21}, where extreme values of $\tauf=9.28$ were reported. Figure~\ref{fig:L63_validation} shows a typical example of $x(t)$ for a forecast time of $\tauf=2.6$, which is close to the mean forecast time. Figure~\ref{fig:L63_attractor} shows that the surrogate model (\ref{eq:Wr}) obtained using RAFDA produces a model which very well approximates the long-time statistics of the full Lorenz system (\ref{eq:L63}) in the sense that its trajectories reproduce the attractor in delay-coordinate space. Contrary, LR does not lead to surrogate models which are consistent with the dynamics of the Lorenz system (\ref{eq:L63}). Reproducing dynamically consistent trajectories is, of course, paramount for the purpose of using such surrogate models for long-time integration when the interest is rather on the overall statistical behaviour rather than on accurate short-term forecasts.    

\begin{figure}[tbp]
\centering
\includegraphics[width=0.8\columnwidth]{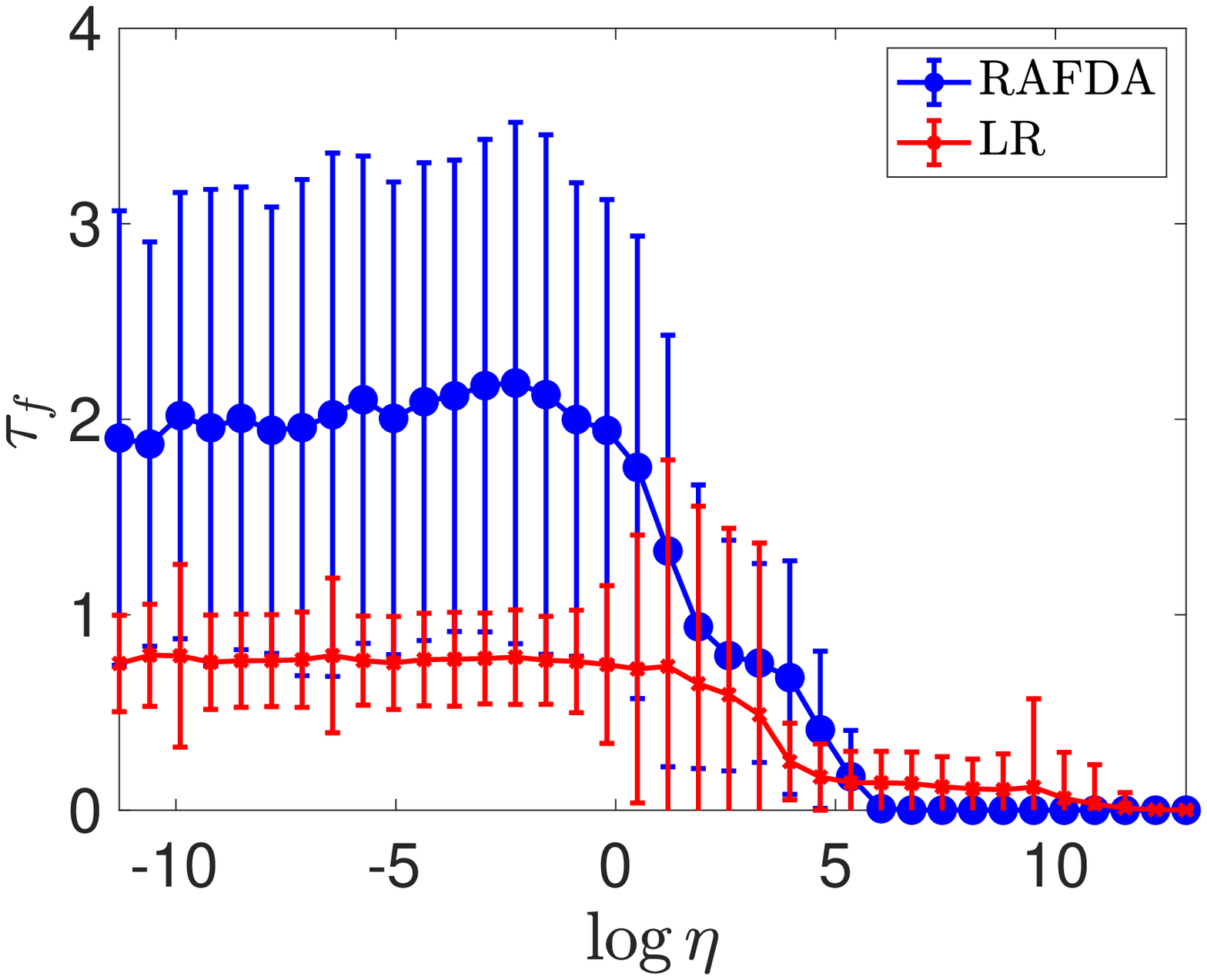}
\caption{Forecast time $\tauf$ as a function of the observational noise strength $\eta$. The error bars denote two standard deviations, estimated from $500$ independent realisations, differing in their randomly drawn internal parameters ($\Win$, $\bin$), training and validation data sets, and in the initial ensembles.}
\label{fig:L63_etaLoop}
\end{figure}

The ability of RAFDA to control noise present in a training data set depends on the strength $\eta$ of the noise. We show in Figure~\ref{fig:L63_etaLoop} results of the mean forecast  time for a range of $\eta \in[6.25 \times 10^{-6},4.3\times 10^5]$. RAFDA outperforms standard random feature maps with linear regression for noise levels $\log \eta<5$. Moreover, there is a robust plateau with mean forecast times of around $\tauf\approx 2$ for a large range of noise strengths with $\eta\lessapprox 1$. Notice that for $\eta\to 0$, maybe surprisingly, the mean forecast time of RAFDA does not converge to the one obtained by standard random feature maps but remains threefold larger. This was discussed in \citet{GottwaldReich21}: although LR achieves a smaller value of the cost function (\ref{eq:RR}) for $\eta \ll 1$ it does not generalise as well to unseen data. The cost function is not zero as generically the output data do not lie in the span of the random feature maps. It is important to note that LR with $\beta>0$ assumes model error (which is not present in our application of simulating the Lorenz-63 system (\ref{eq:L63}) rather than observational error. We further note that, for very large noise levels $\log \eta>5$, the forecast times $\tau_f$ become zero for RAFDA. This is due to the data assimilation component of our method and is an instance of the well-documented problem of filter divergence \citep{Ehrendorfer07,NadigaEtAl13}. More precisely, in finite-size ensembles, most ensemble members may align with the most unstable direction \citep{NgEtAl11} implying a small forecast error covariance. In combination with large observation noise this leads to the filter trusting its own forecast and the analysis is not corrected by incoming new observations. Filter divergence can be avoided by increasing the ensemble size and/or employing covariance inflation. We remark that we abstained here from fine-tuning all hyperparameters.
 
\section{Discussion}
\label{sec:discussion}
We proposed a new data-driven method to estimate surrogate one-step forecast maps from noisy partial observations. Our method determines the surrogate map in the reconstructed phase space for delay vector coordinates, thereby dealing with the problem of having only access to partial observations.  We employed the RAFDA framework in the reconstructed phase space, in which the surrogate map is constructed as a linear combination of random feature maps the coefficients of which are learned sequentially using a stochastic EnKF. We showed that learning the coefficients of the linear combination of random feature maps sequentially with incoming new observations rather than by learning them using the method of least-squares on the whole data set greatly increases the forecast capability of the surrogate model and significantly controls the measurement noise.

Our numerical results showed that the quality of the surrogate model exhibits some degree of variance, depending on the draw of the arbitrarily fixed internal parameters of the random feature map model $\Win$ and $\bin$. Being able to choose good candidates for those parameters or learning them in conjunction with the rest of the surrogate model would greatly improve the applicability of the method. In our previous work we provided some guidance on what constitutes good hyperparameters enabling good learning \citep{GottwaldReich21}. How to choose actual optimal values for these hyperparameters requires, for example, an optimization procedure wrapped around the RAFDA method described here, and is planned for future research. Apart from $\Win$ and $\bin$ there are numerous hyperparameters which require tuning, such as the regularization parameter $\beta$, the inflation parameter $\alpha$, the size and the variance of the initial ensemble, as well as the assumed observational error covariance $\Gamma$. To obtain optimal performance of RAFD these would need to be tuned which can be done by additional optimization procedures  \citep{NadigaEtAl19}.

The random feature map architecture is closely related to reservoir computers \citep{Jaeger02,JaegerHaas04,PathakEtAl18,JuenglingEtAl19}. In reservoir computers the incoming signal is used in conjunction with an internal reservoir dynamics to produce the output. It is argued that this helps to take into account non-trivial memory of the underlying dynamical system. It will be interesting to see if random feature maps in the space of delay vectors performs as well as reservoir computers. For systems with a fast decay of correlation such as the Lorenz 63 system the additional reservoir dynamics did not lead to an improvement of the surrogate map \citep{GottwaldReich21}, but incorporating memory for systems with slow decay of correlation taking into account information from past observations may be important.     


\section*{Acknowledgments}
SR is supported by Deutsche Forschungsgemeinschaft (DFG) - Project-ID 318763901 - SFB1294





\bibliographystyle{abbrvnat}

\end{document}